\documentclass[11pt,a4paper]{article}
\usepackage[hyperref]{emnlp2020}
\usepackage{times}
\usepackage{latexsym}

\usepackage{microtype}

\usepackage{listings}
\usepackage{caption}
\usepackage{subcaption}
\usepackage{graphicx}
\usepackage{multirow}
\usepackage{amsmath}
\usepackage{booktabs}
\usepackage{amsfonts}
\usepackage{balance}
\usepackage{titlesec}
\usepackage{csquotes}

\usepackage{dialogue}
\usepackage{tcolorbox}
\usepackage{mdframed}

\usepackage{blindtext}

\aclfinalcopy 
\def\aclpaperid{3004} 

\usepackage{caption}
\title{Effects of Naturalistic Variation in Goal-Oriented Dialog}

\author{Jatin Ganhotra \\
  IBM Research \\
  \texttt{jatinganhotra@us.ibm.com} \\\And
  Robert Moore \\
  IBM Research \\
  \texttt{rjmoore@us.ibm.com} \\\AND
  Sachindra Joshi \\
  IBM Research \\
  \texttt{jsachind@in.ibm.com} \\\And
  Kahini Wadhawan \\
  IBM Research \\
  \texttt{kawadhaw@in.ibm.com} \\}

\date{}

\begin{document}

\maketitle
\begin{abstract}


Existing benchmarks used to evaluate the performance of end-to-end neural dialog systems lack a key component: natural variation present in human conversations.  
Most datasets are constructed through crowdsourcing, where the crowd workers
follow a fixed template of instructions while enacting the role of a user/agent. 
This results in straight-forward, somewhat routine, and mostly trouble-free conversations, as crowd workers do not think to represent the full range of actions that occur naturally with real users. 
In this work, we investigate the impact of naturalistic variation on two goal-oriented datasets: bAbI dialog task and Stanford Multi-Domain Dataset (SMD). We also propose new and more effective testbeds for both datasets, by introducing naturalistic variation by the user\footnote{The updated test sets are available at: \url{https://github.com/IBM/naturalistic-variation-goal-oriented-dialog-datasets}}. 
We observe that there is a significant drop in performance (more than 60\% in Ent. F1 on SMD and 85\% in per-dialog accuracy on bAbI task) of recent state-of-the-art end-to-end neural methods such as BossNet and GLMP on both datasets.

\end{abstract}

\section{Introduction}
End-to-end dialog systems that learn from human-to-human conversations have huge potential for various goal-oriented dialog tasks such as hotel, restaurant and flight reservations. 
Recent work \cite{serban2016building, bordes2016learning} has shown that it is possible to train dialog models in an end-to-end manner and achieve satisfactory results. There are several benchmarks \cite{wen2017network, el2017frames,eric2017key,wei2018airdialogue} to evaluate the performance of neural models for goal-oriented dialog.

\begin{figure}[t]
\centering
\includegraphics[scale=0.42]{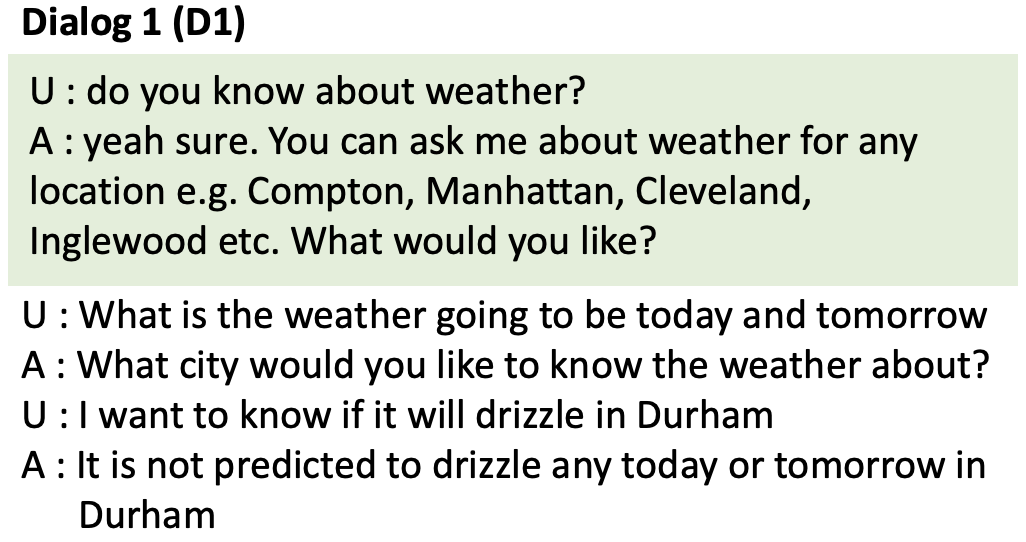}
\caption{ 
Sample dialog from SMD dataset between the User (\textbf{U}) and In-Car Assistant (\textbf{A}). The naturalistic variation added is shown in the box (in green color).
}
\label{fig_dataset}
\end{figure}

However, these benchmarks assume a world of a "perfect" user who always provides precise, concise, and correct utterances. These goal-oriented datasets are largely collected by crowdsourcing, where a crowdsource worker enacts the part of a real user by following a set template of instructions provided for the task. This method results in a dataset where most of the user utterances are straight-forward, stick to the goal and tend to leave out the variation commonly found in naturally occurring conversational data.
For example, in making a restaurant reservation, a user may perform the following actions: a) check on the customer care agent's welfare, b) comment on the weather in the opening of the conversation, c) ask about business hours or about whether the restaurant accepts reservations as a preliminary question to the reservation request and d) paraphrase his/her prior request with more details. 
Each of these actions is natural variation present in human-to-human conversations.

Although some templates ask the crowd workers to paraphrase their request, they never ask workers to simulate the full range of naturalistic variation
\cite{schegloff1977preference, 10.1145/3304087}. This naturalistic variation has been thoroughly documented in the Conversation Analysis literature \cite{10.2307/412243, schegloff_2007}, and further adapted for designing automated conversational agents \cite{10.1145/3304087}.

The core reason for this omission is that naturalistic variation is often confused with "\textit{chit chat}" \cite{dunbar1997human, zhang-etal-2018-personalizing}. 
\citet[p.~121]{10.1145/3304087} writes, 
\begin{displayquote}
\emph{In common usage, “chit chat” means inconsequential talk. But much talk that may appear on the surface to be inconsequential in fact serves a variety of functions in managing the conversation itself.}
\end{displayquote}
In this work, we focus on the full range of activities observed in naturally occurring conversations, referred to as "\textit{natural variation}".
Our goal in this work is three-fold:
\begin{itemize}
    \setlength\itemsep{-0.2em}
    \item Highlight the problem of unnatural data generated through crowdsourcing
    \item Showcase the impact of natural variation in the performance of state-of-the-art dialog systems, and
    \item Publicly release improved testbeds for two datasets used extensively in goal-oriented dialog research: bAbI dialog task and SMD.
\end{itemize}{}

Recently, few approaches have been explored to study the behavior of neural dialog systems in the presence of synthetically introduced perturbations to the dialog history. \citet{eshghi2017bootstrapping} created the bAbI+ dataset, an extension of bAbI dialog task-1, by introducing variations like hesitations, restarts and corrections.
\citet{zhao2018zero} created SimDial, which simulates spoken language phenomena, e.g. self-repair and hesitation. \citet{DBLP:conf/acl/SankarSPCB19} introduce utterance-level and word-level perturbations on various benchmarks. 
However, such variations have been largely artificial and do not reflect the "\emph{natural variation}" commonly found in naturally occuring conversational data.

\citet{geva2019we} show that often models do not generalize well to examples from new annotators at test time who did not contribute to training data, which reinforces our choice of introducing natural variation in the test set for evaluation.


\section{Datasets}
\label{datasets}


We study and observe issues in multiple goal-oriented dialog benchmarks. In this work, we focus on two multi-turn goal-oriented datasets: bAbI dialog task and SMD for evaluating the impact of natural variation. We provide details on issues in the following datasets: SMD, CamRest676 \cite{wen2017network}, Frames \cite{el2017frames} and AirDialogue \cite{wei2018airdialogue} in the Appendix.

\subsection{bAbI dialog task}
\label{bAbI-dialog-task-5-dataset}

The bAbI dialog tasks dataset \cite{bordes2016learning} includes five simulated tasks in the restaurant domain, where the dialog system has to retrieve the correct response from a set of given candidate responses.
Task 1 to 4 are sub-tasks about issuing and updating API calls, recommending restaurant options, and providing additional information about a restaurant. Task 5 combines all tasks. 
There are two KBs used, where one KB is used to generate the standard training, validation, and test sets, and the other KB is used only to generate an Out-Of-Vocabulary (OOV) test set.
The task is considered simple due to the small number of user and agent responses but is used extensively for goal-oriented dialog research. 


\subsection{Stanford Multi-Domain dataset (SMD)}
\begin{figure}[t]
\centering
\includegraphics[scale=0.4]{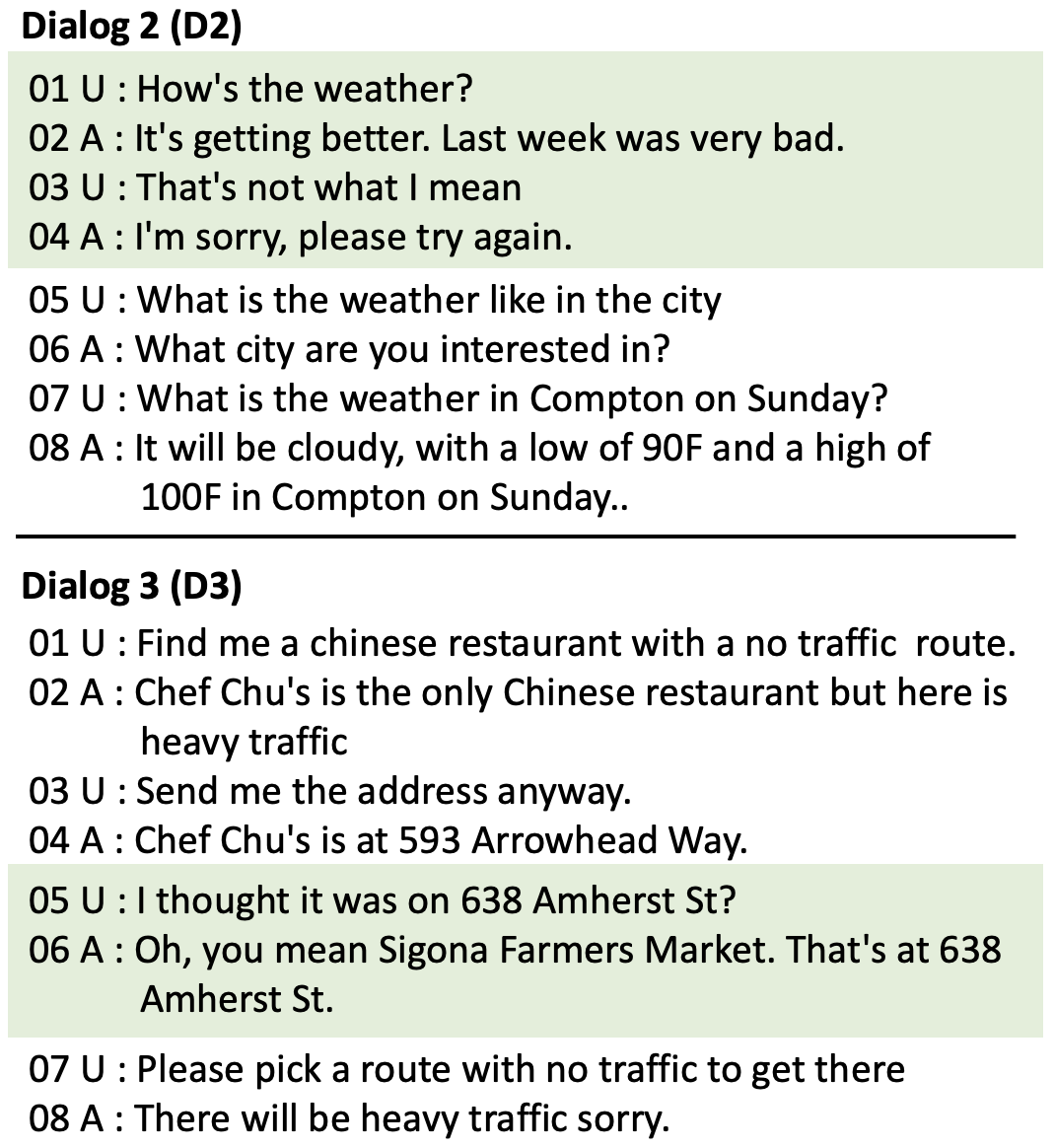}
\caption{ 
Additional dialogs from SMD dataset between the User (U) and In-Car Assistant (A). The natural variation \textbf{NCF} pattern added to \textbf{D2}: \emph{(B) Misunderstanding Report} and \textbf{D3}: \emph{(B) Other Correction} are shown in boxes (in green color). 
}
\label{fig_dataset_2_examples}
\end{figure}

SMD \cite{eric2017key} is a multi-domain, task-oriented dialog dataset with three distinct domains: calendar scheduling, weather information retrieval, and point-of-interest navigation.  
SMD was collected using a Wizard-of-Oz (Woz) approach inspired by \citet{wen2017network}. 
We provide sample dialogs in Figure \ref{fig_dataset_2_examples}. Crowd workers had two roles: Driver (tasked to extract certain information from the Car Assistant) and Car Assistant (tasked to answer Driver query using a private KB). 


We incorporate the naturalistic variation mentioned below in Section \ref{naturalistic-conversations} to these datasets as they are used extensively in goal-oriented dialog research and use them as benchmarks for our experimental evaluation\footnote{The updated test sets for bAbI dialog task-5 and SMD are available at: \url{https://github.com/IBM/naturalistic-variation-goal-oriented-dialog-datasets}}. Note that we introduce variation only in the test sets and create additional updated-test sets to simulate the presence of natural variation during deployment.


\section{Naturalistic variation}
\label{naturalistic-conversations}

In order to better approximate natural variation in our datasets, we utilize the \emph{Natural Conversation Framework} (\textbf{NCF}) \cite{10.1145/3304087}. The NCF is a framework for designing conversational agents that more closely emulate natural conversation than most of today's chatbots and voice assistants. 
The NCF is organized into two kinds of patterns: \emph{conversational activities} and \emph{conversation management}. The conversational activity patterns (denoted by $A$) handle the main business of conversation, i.e.\ the user request and the services provided by agent.
On the other hand, conversation management patterns help the user and agent to manage the conversation itself. Conversation management occurs on two levels: sequence level (denoted $B$) and conversation level (denoted $C$).

\begin{table}[t]
\begin{center}
\begin{tabular}{l|cc}
    \hline
    \textbf{Pattern} & \textbf{SMD} & \textbf{bAbI}\\ \hline
    \small (A) Open Request Screening & 64 & 54\\ \hline
    \small (A) Open Request User Detail Request & - & 143 \\ \hline
    \small (B) Example Request & 23 & -\\ \hline
    \small (B) Misunderstanding Report & 35 & 314\\ \hline
    \small (B) Other Correction & 24 & 522\\ \hline
    \small (B) Sequence Closer (not helped) & 6 & 811\\ \hline
    \small (B) Sequence Closer (repaired) & 139 & 189\\ \hline
    \small (C) Capability Expansion & 151 & 811\\ \hline
    \small (C) Recipient Correction & 100 & -\\ \hline
    
\end{tabular}
\end{center}
\caption{\# of dialogs updated per pattern out of total \# of dialogs in test set (SMD - 304, bAbI - 1000). The '$-$' entry means that the specific pattern was not added to dialogs in the given dataset.}
\label{tab:dataset-stats}
\end{table}

\begin{table}[t]
\begin{center}
\begin{tabular}{l|cc}
    \hline
    \textbf{\# patterns} & \textbf{SMD} & \textbf{bAbI}\\ \hline
    $1$ & 288 & 1000\\ \hline
    $>1$ & 198 & 981\\ \hline
    $>2$ & 57 & 843\\ \hline
    $>3$ & 7 & 375\\ \hline
    $>4$ & 0 & 4\\ \hline
\end{tabular}
\end{center}
\caption{\# of dialogs updated per \# of patterns out of total \# of dialogs in test set (SMD - 304, bAbI - 1000).}
\label{tab:dataset-stats-num-patterns}
\end{table}

After studying the 100 patterns in NCF, we identified a subset of the 32 patterns which are most commonly found in goal-oriented natural conversations 
and use them in our work. We excluded the remaining 68 patterns that covered other types of conversations e.g. quiz, question-answer jokes, voice-based assistant setting, etc. We provide details of these 32 patterns in the Appendix.
For each pattern $p$ from the 32 NCF patterns, we identify conversations in the test set where $p$ could have been present in the conversation if the crowd worker was not limited to a given template. We define rules and heuristics based on the annotations 
e.g.\ dialog acts, slot information, etc. captured by the crowdsource worker from the user utterance. After introducing additional user utterances and agent responses per NCF pattern, we perform manual review 
of 20\% of updated dialogs randomly to ensure that incorporating the pattern does not make the dialog incoherent \cite{DBLP:conf/acl/SankarSPCB19}. After manual review, 
we select a subset of 9 patterns 
and incorporate them in the two datasets.

The statistics for number of dialogs in the test sets for both datasets updated per pattern are in Table \ref{tab:dataset-stats}, and Table \ref{tab:dataset-stats-num-patterns} provides details on number of dialogs where more than 1 pattern was added. We provide details with examples for a few patterns below and share details for the rest in the Appendix. Each pattern is denoted as pattern class ($A/B/C$) followed by \emph{pattern type}.

($A$) \emph{Open Request Screening}: The user asks a preliminary question to a complex request to determine if the agent may be able to help with it. e.g.\ 
dialog \textit{D1} in Figure \ref{fig_dataset}.

($B$) \emph{Misunderstanding Report}: The user tells the agent that it misunderstood what he or she said, e.g.\ 
(line 03) in dialog \textit{D2} in Figure \ref{fig_dataset_2_examples}.

($C$) \emph{Capability Expansion}: The user asks the agent to expand on one of its own capabilities that it previously mentioned, e.g.\ "Tell me more about restaurant recommendations."


Although the naturalistic variation increases the complexity of the dialog, the added utterances do not increase the complexity of the goal, in other words, they do not introduce new topics or courses of action, they merely expand the existing ones.


\begin{table}[t]
\begin{center}
\begin{tabular}{l|cc}
    \hline
    \textbf{Model} & \textbf{BLEU} & \textbf{Ent. F1}\\ \hline
    
    Bossnet test & 5.42 & 36.34 \\ \hline
    Bossnet test-updated & 3.7 & 21.81 \\ \hline
    \hline
    GLMP test & 14.22 & 55.38\\ \hline
    GLMP test-updated & 4.73 & 21.05 \\ \hline

\end{tabular}
\end{center}
\caption{Performance of models on (original and updated) test sets for SMD dataset}
\label{tab:baseline-results-smd}
\end{table}

\begin{table}[t]
\begin{center}
\begin{tabular}{l|l||l}
    \hline
    \textbf{Task/Model} & \textbf{Bossnet} & \textbf{GLMP}\\ \hline
    T5 & 97.82 (67.2) & 99.20 (88.5) \\ \hline
    T5-updated & 90.4 (37.9) & 87.24 (12.7) \\ \hline
    T5-OOV & 90.77 (12.1) & 92.33  (21.8) \\ \hline
    T5-OOV-updated & 83.65 (7.0) & 83.97 (5.9) \\ \hline
\end{tabular}
\end{center}
\caption{Per-response (per-dialog) accuracy of models on (original and updated) test and test-OOV sets for bAbI dialog task-5 (denoted as T5 above)}
\label{tab:baseline-results-bAbI-new}
\end{table}

\section{Experiments}
\label{experiments}
We use two state-of-the-art models: \textit{BossNet} \cite{raghu2019disentangling} and \textit{GLMP} \cite{DBLP:conf/iclr/WuSX19} as the baselines for our experiments. We use the best performing hyper-parameters reported by both models for each dataset. The test results reported (in Table \ref{tab:baseline-results-smd} and \ref{tab:baseline-results-bAbI-new}) are calculated by using the saved model with highest validation performance across multiple runs. Training setting and hyperparameter details for both models in Appendix. 

For evaluation of the synthetic bAbI dialog task-5, we use per-response and per-dialog accuracy \cite{bordes2016learning}. For SMD, we use a) BLEU \cite{papineni2002bleu} and b) Entity F1 \cite{eric2017key} scores.
We evaluate BossNet and GLMP models on both the original and the updated test set. We do not evaluate the models on their ability to generate the newly added system responses as part of the naturalistic variation, but only on the system responses originally present in the test set.

\begin{table}[t]
\begin{center}
\begin{tabular}{l|cc}
    \hline
    \textbf{Pattern} & \textbf{BLEU} & \textbf{Ent F1} \\ \hline
    
    \small Original test set & 14.22 & 55.38 \\ \hline
    \small Updated test set & 4.73 & 21.05 \\ \hline
    \small (A) Open Request Screening & 11.63 & 51.61 \\ \hline
    \small (B) Example Request & 14.23 & 53.51 \\ \hline
    \small (B) Misunderstanding Report & 12.91 & 55.01 \\ \hline
    \small (B) Other Correction & 14.1 & 55.21 \\ \hline
    \small (B) Sequence Closer (not helped) & 14.24 & 55.23 \\ \hline
    \small (B) Sequence Closer (repaired) & 14.69 & 55.20 \\ \hline
    \small (C) Capability Expansion & 8.29 & 27.61 \\ \hline
    \small (C) Recipient Correction & 13.01 & 50.22 \\ \hline
    
\end{tabular}
\end{center}
\caption{Ablation results for GLMP model on SMD}
\label{tab:ablation-study}
\end{table}

\section{Results}
\label{results}

From Table \ref{tab:baseline-results-smd} and \ref{tab:baseline-results-bAbI-new}, we observe that both models perform very poorly on our updated-test sets. For SMD, the EntF1 score drops by 62\% for GLMP and 40\% for BossNet. We observe similar performance reduction trends for bAbI dialog task-5,  where the per-dialog accuracy decreases by more than 43\% for BossNet and 85\% for GLMP model.

We observe that the drop in performance on bAbI is much less than SMD. This is because bAbI is a synthetic dataset with a small set of fixed agent responses. Since the models are evaluated only on the agent responses present in the original test set, additional user and agent utterances for incorporating natural variation do not affect performance too much. On the other hand, SMD is a real-world dataset of human-to-human conversations collected by crowdsourcing and we observe a much higher drop across both BLEU and Ent F1 scores.

We perform ablation experiments to study the impact of each pattern (presented in Table \ref{tab:ablation-study}). We create separate updated-test sets for SMD for each pattern, by adding only one pattern at a time for the same number of dialogs per pattern from Table \ref{tab:dataset-stats}. We observe that ($C$) \emph{Capability Expansion} pattern hurts the GLMP model performance the most in comparison to other patterns. 
As mentioned in Sec \ref{naturalistic-conversations}, in Capability Expansion, the user asks details from the agent about its capabilities. Since SMD has three domains, this adds more user and agent utterances to the dialog history, in comparison to other patterns, which results in a larger drop in model performance. In addition to higher overall dialog length, new domain entities are also present in these new utterances where agent/bot provides details on the services available, which results in lower performance. We provide statistics for change in average number of utterances per dialog per pattern for SMD in the Appendix.



Our results clearly show that naturalistic variation present during deployment affects model performance and will result in lower than expected performance for a given dialog system in production.



\section{Conclusion}
\label{conclusion}
This work studies the dangers of using crowd-sourced data, without templates for the natural range of activities in conversation, such as the Natural Conversation Framework \cite{10.1145/3304087}, to train end-to-end dialog systems. 
We highlight the impact on the performance of state-of-the-art models on our new and effective testbeds for bAbI dialog task-5 and SMD datasets, which have naturalistic variation.
We believe this opens up a new and promising research direction for devising improved strategies for crowdsourcing goal-oriented datasets, as well as improved models that can better handle interactions with real users.


\bibliographystyle{acl_natbib}
\bibliography{anthology,emnlp2020}

\begin{thebibliography}{20}
\expandafter\ifx\csname natexlab\endcsname\relax\def\natexlab#1{#1}\fi

\bibitem[{Bordes et~al.(2017)Bordes, Boureau, and Weston}]{bordes2016learning}
Antoine Bordes, Y-Lan Boureau, and Jason Weston. 2017.
\newblock Learning end-to-end goal-oriented dialog.
\newblock \emph{In the International Conference on Learning Representations
  (ICLR)}.

\bibitem[{Dunbar et~al.(1997)Dunbar, Marriott, and Duncan}]{dunbar1997human}
Robin~IM Dunbar, Anna Marriott, and Neil~DC Duncan. 1997.
\newblock Human conversational behavior.
\newblock \emph{Human nature}, 8(3):231--246.

\bibitem[{El~Asri et~al.(2017)El~Asri, Schulz, Sharma, Zumer, Harris, Fine,
  Mehrotra, and Suleman}]{el2017frames}
Layla El~Asri, Hannes Schulz, Shikhar Sharma, Jeremie Zumer, Justin Harris,
  Emery Fine, Rahul Mehrotra, and Kaheer Suleman. 2017.
\newblock Frames: a corpus for adding memory to goal-oriented dialogue systems.
\newblock In \emph{Proceedings of the 18th Annual SIGdial Meeting on Discourse
  and Dialogue}, pages 207--219.

\bibitem[{Eric and Manning(2017)}]{eric2017key}
Mihail Eric and Christopher~D Manning. 2017.
\newblock Key-value retrieval networks for task-oriented dialogue.
\newblock \emph{In Proceedings of the 18th Annual SIGdial Meeting on Discourse
  and Dialogue (SIGDIAL)}.

\bibitem[{Eshghi et~al.(2017)Eshghi, Shalyminov, and
  Lemon}]{eshghi2017bootstrapping}
Arash Eshghi, Igor Shalyminov, and Oliver Lemon. 2017.
\newblock Bootstrapping incremental dialogue systems from minimal data: the
  generalisation power of dialogue grammars.
\newblock In \emph{Proceedings of the 2017 Conference on Empirical Methods in
  Natural Language Processing}, pages 2220--2230.

\bibitem[{Geva et~al.(2019)Geva, Goldberg, and Berant}]{geva2019we}
Mor Geva, Yoav Goldberg, and Jonathan Berant. 2019.
\newblock Are we modeling the task or the annotator? an investigation of
  annotator bias in natural language understanding datasets.
\newblock \emph{arXiv preprint arXiv:1908.07898}.

\bibitem[{Kelley(1984)}]{kelley1984iterative}
John~F Kelley. 1984.
\newblock An iterative design methodology for user-friendly natural language
  office information applications.
\newblock \emph{ACM Transactions on Information Systems (TOIS)}, 2(1):26--41.

\bibitem[{Moore and Arar(2019)}]{10.1145/3304087}
Robert~J. Moore and Raphael Arar. 2019.
\newblock \emph{Conversational UX Design: A Practitioner’s Guide to the
  Natural Conversation Framework}.
\newblock Association for Computing Machinery, New York, NY, USA.

\bibitem[{Papineni et~al.(2002)Papineni, Roukos, Ward, and
  Zhu}]{papineni2002bleu}
Kishore Papineni, Salim Roukos, Todd Ward, and Wei-Jing Zhu. 2002.
\newblock Bleu: a method for automatic evaluation of machine translation.
\newblock In \emph{Proceedings of the 40th annual meeting on association for
  computational linguistics}, pages 311--318. Association for Computational
  Linguistics.

\bibitem[{Raghu et~al.(2019)Raghu, Gupta et~al.}]{raghu2019disentangling}
Dinesh Raghu, Nikhil Gupta, et~al. 2019.
\newblock Disentangling language and knowledge in task-oriented dialogs.
\newblock In \emph{Proceedings of the 2019 Conference of the North American
  Chapter of the Association for Computational Linguistics: Human Language
  Technologies, Volume 1 (Long and Short Papers)}, pages 1239--1255.

\bibitem[{Sacks et~al.(1974)Sacks, Schegloff, and Jefferson}]{10.2307/412243}
Harvey Sacks, Emanuel~A. Schegloff, and Gail Jefferson. 1974.
\newblock \href {http://www.jstor.org/stable/412243} {A simplest systematics
  for the organization of turn-taking for conversation}.
\newblock \emph{Language}, 50(4):696--735.

\bibitem[{Sankar et~al.(2019)Sankar, Subramanian, Pal, Chandar, and
  Bengio}]{DBLP:conf/acl/SankarSPCB19}
Chinnadhurai Sankar, Sandeep Subramanian, Chris Pal, Sarath Chandar, and Yoshua
  Bengio. 2019.
\newblock \href {https://www.aclweb.org/anthology/P19-1004/} {Do neural dialog
  systems use the conversation history effectively? an empirical study}.
\newblock In \emph{Proceedings of the 57th Conference of the Association for
  Computational Linguistics, {ACL} 2019, Florence, Italy, July 28- August 2,
  2019, Volume 1: Long Papers}, pages 32--37. Association for Computational
  Linguistics.

\bibitem[{Schegloff(2007)}]{schegloff_2007}
Emanuel~A. Schegloff. 2007.
\newblock \href {https://doi.org/10.1017/CBO9780511791208} {\emph{Sequence
  Organization in Interaction: A Primer in Conversation Analysis}}, volume~1.
\newblock Cambridge University Press.

\bibitem[{Schegloff et~al.(1977)Schegloff, Jefferson, and
  Sacks}]{schegloff1977preference}
Emanuel~A Schegloff, Gail Jefferson, and Harvey Sacks. 1977.
\newblock The preference for self-correction in the organization of repair in
  conversation.
\newblock \emph{Language}, 53(2):361--382.

\bibitem[{Serban et~al.(2016)Serban, Sordoni, Bengio, Courville, and
  Pineau}]{serban2016building}
Iulian~Vlad Serban, Alessandro Sordoni, Yoshua Bengio, Aaron~C Courville, and
  Joelle Pineau. 2016.
\newblock Building end-to-end dialogue systems using generative hierarchical
  neural network models.
\newblock In \emph{AAAI}, volume~16, pages 3776--3784.

\bibitem[{Wei et~al.(2018)Wei, Le, Dai, and Li}]{wei2018airdialogue}
Wei Wei, Quoc Le, Andrew Dai, and Jia Li. 2018.
\newblock Airdialogue: An environment for goal-oriented dialogue research.
\newblock In \emph{Proceedings of the 2018 Conference on Empirical Methods in
  Natural Language Processing}, pages 3844--3854.

\bibitem[{Wen et~al.(2017)Wen, Vandyke, Mrk{\v{s}}i{\'c}, Gasic, Barahona, Su,
  Ultes, and Young}]{wen2017network}
Tsung-Hsien Wen, David Vandyke, Nikola Mrk{\v{s}}i{\'c}, Milica Gasic, Lina
  M~Rojas Barahona, Pei-Hao Su, Stefan Ultes, and Steve Young. 2017.
\newblock A network-based end-to-end trainable task-oriented dialogue system.
\newblock In \emph{Proceedings of the 15th Conference of the European Chapter
  of the Association for Computational Linguistics: Volume 1, Long Papers},
  pages 438--449.

\bibitem[{Wu et~al.(2019)Wu, Socher, and Xiong}]{DBLP:conf/iclr/WuSX19}
Chien{-}Sheng Wu, Richard Socher, and Caiming Xiong. 2019.
\newblock \href {https://openreview.net/forum?id=ryxnHhRqFm} {Global-to-local
  memory pointer networks for task-oriented dialogue}.
\newblock In \emph{7th International Conference on Learning Representations,
  {ICLR} 2019, New Orleans, LA, USA, May 6-9, 2019}. OpenReview.net.

\bibitem[{Zhang et~al.(2018)Zhang, Dinan, Urbanek, Szlam, Kiela, and
  Weston}]{zhang-etal-2018-personalizing}
Saizheng Zhang, Emily Dinan, Jack Urbanek, Arthur Szlam, Douwe Kiela, and Jason
  Weston. 2018.
\newblock \href {https://doi.org/10.18653/v1/P18-1205} {Personalizing dialogue
  agents: {I} have a dog, do you have pets too?}
\newblock In \emph{Proceedings of the 56th Annual Meeting of the Association
  for Computational Linguistics (Volume 1: Long Papers)}, pages 2204--2213,
  Melbourne, Australia. Association for Computational Linguistics.

\bibitem[{Zhao and Eskenazi(2018)}]{zhao2018zero}
Tiancheng Zhao and Maxine Eskenazi. 2018.
\newblock Zero-shot dialog generation with cross-domain latent actions.
\newblock In \emph{Proceedings of the 19th Annual SIGdial Meeting on Discourse
  and Dialogue}, pages 1--10.

\end{thebibliography}


\vspace{0.3cm}

\appendix


\section{Appendix: Natural Conversation Framework (NCF) patterns}
\label{appendix_patterns}

At the core of the NCF is a pattern language of 100 interaction patterns that are adapted from conversation science for modeling rule-based dialog. NCF pattern language is organized into three classes: A) conversational activity, B) sequence-level management, and C) conversation-level management.

The conversational activity patterns (A) involve the main business of the interaction and include ways in which user or agent can request information from the other (A1, A5), ways in which users can make complex requests in an open-ended way (A2), ways in which agents can tell stories or give instructions interactively (A3), and ways in which agents can quiz users on any topics (A4).

The sequence-level management patterns (B) involve managing particular sequences of utterances and include ways in which the agent and the user can repair troubles in hearing or understanding immediately prior utterances (B1, B2) or earlier utterances (B3), as well as ways of ending sequences either by closing them (B4) or by aborting them (B5).

Finally, the conversation-level management patterns (C) involve coordinating entry into and exit from the interaction itself and include ways in which agents or the user can open the conversation (C1, C2), ways they can talk about the agent’s capabilities (C3), and ways they can end the conversation either by closing it (C4) or disengaging from each other in other ways.

Each pattern consists of an abstract model in the form of a transcript with generic social actions. For example, Pattern A2.3 - Open Request is described below (Listing \ref{pattern-a-2-3}). The line numbers refer to utterance number in the conversation, U and A refer to user and agent utterance and generic social actions are listed in capitals. 

\begin{lstlisting}[caption=Pattern A2.3 - Open Request Screening,captionpos=b,label={pattern-a-2-3}]
1 U: PRE-REQUEST
2 A: GO-AHEAD
3 U: FULL REQUEST
4 A: GRANT
5 U: SEQUENCE CLOSER
6 A: RECEIPT
\end{lstlisting}

We provide details on other NCF patterns which were incorporated in the datasets, but omitted in the main paper due to space limitations below;
\begin{itemize}
    \item A: Open Request User Detail Request is a pattern in which the user requests additional information when attempting to answer an agent question, for example, "What are my choices?"
    \item B: Other Correction is a pattern in which the agent corrects the user's second to last utterance based on his or her last utterance, for example, "Oh, you mean a different place."
    \item B: Sequence Closer Not Helped is a pattern in which the user acknowledges a response from the agent in a negative way when it was not helpful, for example, "too bad" or "oh well."
    \item B: Sequence Closer Repaired is a pattern in which the user acknowledges the repair of a part of a sequence, for example, an "ok" or "thank you" after the agent provides a repeat, paraphrase, example, etc.
    \item B: Example Request is a pattern in which the user requests clarification of the agent's prior utterance in the form of an example, for example, "Can you give an example?"
    \item C: Recipient Correction is a pattern in which the user indicates that he or she is talking to someone other than the agent, for example, "I'm not talking to you."
\end{itemize}

\section{Appendix: NCF patterns for goal-oriented dialog}

We provide the list of 32 patterns from the 100 NCF patterns, which are most commonly found in goal-oriented natural conversations below:

\textbf{(A): Conversational Activity Patterns}
\begin{itemize}
    \setlength\itemsep{-0.2em}
    \item A1.1 Inquiry (User) Confirmation
    \item A1.2 Inquiry (User) Disconfirmation
    \item A1.3 Inquiry (User) Repairs
    \item A2.2. Open Request Continuer
    \item A2.3 Open Request Screening
    \item A2.5 Open Request User Detail Request
    \item A2.6 Open Request Summary
    \item A2.11 Open Request Repairs
    \item A3.0 Extended Telling with Repair
    \item A3.1 Extended Telling Abort
\end{itemize}{}

\textbf{(B): Sequence-Level Management Patterns}
\begin{itemize}
    \setlength\itemsep{-0.2em}
    \item B1.2.2 Agent Continuer
    \item B2.6.0 Example Request
    \item B3.1.1 Misunderstanding Report
    \item B3.2.0 Other-Correction
    \item B4.0 Sequence Closer (helped)
    \item B4.1 Sequence Closer (not helped)
    \item B4.2 Sequence Closer Appreciation
    \item B4.4 Sequence Closer (repaired)
\end{itemize}{}

\textbf{(C): Conversation-Level Management Patterns}
    \begin{itemize}
    \setlength\itemsep{-0.2em}
    \item C1.4 Opening Welfare Check (Agent)
    \item C1.5 Opening Organization Offer of Help (Agent)
    \item C1.7 Organizational Problem Request (Agent)
    \item C2.1 Summons (User)
    \item C2.2 Welfare Check (User)
    \item C2.9 Name Correction (User)
    \item C3.0 General Capability Check
    \item C3.1 Capability Expansion
    \item C3.2 Specific Capability Check
    \item C4.7 Closing Success Check (Disaffirmed)
    \item C4.8 Closing Success Check Reopened
    \item C4.9 Closing Offer (Affirmed)
    \item C4.10 Closing Offer (Disaffirmed)
    \item C5.2 Recipient Correction

\end{itemize}{}

\section{Appendix: Issues with existing benchmarks}
\label{issues-existing-datasets}

\subsection{SMD}
In the SMD dataset, to encourage diversity in the discourse, some knowledge bases intentionally lacked attributes. To encourage more naturalistic, unbiased utterances, crowd workers (enacting a user) were also asked to record voice commands of actions a car assistant could perform, which were transcribed and used as the first user utterance in a given dialog. However, this technique was limited only to the first-utterance and further only for 50\% of total dialogs. Overall, the majority of the user utterances in the dataset are specific commands, or concise and direct questions about information in KB\footnote{Refer to Appendix - \citet{eric2017key}.}. 

\subsection{CamRest676}
\citet{wen2017network} used the Wizard-of-Oz (WOz) approach \cite{kelley1984iterative} and designed a system to assist users to find a restaurant in the Cambridge, UK area. The setting is similar to the restaurant table booking simulated dataset, dialog bAbI tasks, in \citet{bordes2016learning} and collected 676 human-to-human dialogues. There were three informable slots (food, price range, area) that participants in the user role used to constrain the search (similar to bAbI dialog task-1) and six requestable slots (address, phone, postcode and the three informable slots) that the user could ask about once a restaurant has been offered (similar to bAbI dialog task-4). However, the user utterances in the dataset are straight-forward and always stick to the point without any diversity and novelty in natural language\footnote{Refer to Appendix: Sample dialogues - \citet{wen2017network}}.

\subsection{Frames}
\citet{el2017frames} presented Frames corpus, by also using the Wizard-of-Oz (WOz) approach where the participants in the user role were given task templates during the data collection process. From the 38 templates used, 14 templates were generic and the other 24 were written to encourage more role-playing from users. This resulted in some novelty in the data collected and prevented the user utterances to be repetitive. However, to control data collection, the participants were asked to follow a set of instructions which resulted in user utterances largely focused on the task.

\subsection{AirDialogue}
\citet{wei2018airdialogue} recently presented AirDialogue, a large goal-oriented dataset where human annotators play the role of a customer or an agent and interact with the goal of successfully booking a trip given travel and flight restrictions generated by a context-generator. The dataset is the largest currently as it has largest context complexity and state complexity (based on all possible combinations of customer and agent context features, like number of flights in the database, number of airlines, airport codes and dialogue action states), in comparison to other existing datasets mentioned above. However, the authors don't share details on how the dataset was collected and instructions provided to the participants\footnote{AirDialogue dataset has not been publicly released to the research community.}.

\section{Appendix: Average Utterance count for SMD dialogs per pattern}
As part of our ablation experiments, we create separate updated-test sets for SMD for each pattern, by adding only one pattern at a time for the same number of dialogs per pattern as mentioned in the dataset statistics. For the same dataset, we provide the corresponding matching values for the average utterance count for SMD test set dialogs per pattern in Table \ref{tab:dataset-stats-avg-utt-count}.

\begin{table}[htbp!]
\begin{center}
\begin{tabular}{l|c}
    \hline
    \textbf{Pattern} & 
    \textbf{avg. \# utt.}\\ \hline
    \small (A) Open Request Screening & 5.77 \\ \hline
    \small (B) Example Request & 5.5 \\ \hline
    \small (B) Misunderstanding Report & 5.81\\ \hline
    \small (B) Other Correction & 5.51\\ \hline
    \small (B) Sequence Closer (not helped) & 5.39\\ \hline
    \small (B) Sequence Closer (repaired) & 6.27 \\ \hline
    \small (C) Capability Expansion & 10.32\\ \hline
    \small (C) Recipient Correction & 7.99\\ \hline
    \hline
    \small Original test set & 5.35 \\ \hline
    
\end{tabular}
\end{center}
\caption{Average number of utterances per dialog for the SMD test set after adding each pattern.}
\label{tab:dataset-stats-avg-utt-count}
\end{table}

\section{Appendix: Training Details}
We use the best performing hyper-parameters reported by both models - BossNet and GLMP for each dataset. The test results reported are calculated by using the saved model with highest validation performance across multiple runs. Training setting and hyperparameter details for both models are provided below.

\subsection{Baseline method: Bossnet}
The hyperparameters used to train Bossnet on the different datasets are provided in Table \ref{tab:bossnet-params}.

\begin{table}[ht]
\begin{center}
\begin{tabular}{l|cc}
\toprule
\textbf{Parameter} & \textbf{T5} & \textbf{SMD}\\
\midrule
Learning Rate & 0.0005 & 0.0005 \\ \hline
Hops & 3 & 3 \\ \hline
Embedding Size & 256 & 256  \\ \hline
Disentangle Loss Weight & 1.0 & 1.0 \\ \hline
Disentangle Label Dropout & 0.2 & 0.1 \\ \hline
\end{tabular}
\caption{The hyperparameters used to train Bossnet on the bAbI-dialog-task-5 (denoted T5) and SMD datasets}.
\label{tab:bossnet-params}
\end{center}
\end{table}

\subsection{Baseline method: GLMP}

We use GLMP K3 (hops = 3) for training on the SMD dataset and GLMP K1 (hops = 1) for training on bAbI dialog task-5, as this configuration provides the best results. For both datasets, we used learning rate equal to 0.001, with a decay rate of 0.5. The hyperparameters used to train GLMP on the different datasets are provided in Table \ref{tab:glmp-params}.

\begin{table}[h]
\begin{center}
\begin{tabular}{l|cc}
\toprule
\textbf{Parameter} & \textbf{T5} & \textbf{SMD}\\
\midrule
Hops & 1 & 3 \\ \hline
Embedding dimension & 128 & 128 \\ \hline
GRU hidden size & 128 & 128 \\ \hline
Dropout rate & 0.3 & 0.2 \\ \hline
\end{tabular}
\caption{The hyperparameters used to train GLMP on the bAbI-dialog-task-5 (denoted T5) and SMD datasets.}
\label{tab:glmp-params}
\end{center}
\end{table}

\end{document}